\documentclass{article}

\usepackage{arXiv}

\usepackage[utf8]{inputenc} % allow utf-8 input
\usepackage[T1]{fontenc}    % use 8-bit T1 fonts
\usepackage{hyperref}       % hyperlinks
\usepackage{url}            % simple URL typesetting
\usepackage{booktabs}       % professional-quality tables
\usepackage{amsfonts}       % blackboard math symbols
\usepackage{nicefrac}       % compact symbols for 1/2, etc.
\usepackage{microtype}      % microtypography
\usepackage{xcolor}         % colors
\usepackage{graphicx}
\usepackage{wrapfig}
\usepackage{amsmath}
\usepackage{pifont}
\usepackage{subcaption}
\title{BIPNN: Learning to Solve Binary Integer Programming via Hypergraph Neural Networks}

\author{%
  Sen Bai\\
  Changchun University of Science\\
  and Technology, China \\
  \texttt{baisen@cust.edu.cn} \\
  \And
  Chunqi Yang \\
  Changchun University of Science\\
  and Technology, China \\
  \texttt{yangchunqi@mails.cust.edu.cn} \\
  \AND
  Xin Bai \\
  Huawei Technologies Co. Ltd \\
  China \\
  \texttt{baixinbs@163.com} \\
  \And
  Xin Zhang \\
  Changchun University of Science\\
  and Technology, China \\
  \texttt{zhangxin@cust.edu.cn} \\
  \And
  Zhengang Jiang \\
  Changchun University of Science\\
  and Technology, China \\
  \texttt{jiangzhengang@cust.edu.cn} \\
}

\begin{document}

\maketitle

\begin{abstract}
Binary (0-1) integer programming (BIP) is pivotal in scientific domains requiring discrete decision-making.
As the advance of AI computing, recent works explore neural network-based solvers for integer linear programming (ILP) problems.
Yet, they lack scalability for tackling nonlinear challenges.
To handle nonlinearities, state-of-the-art Branch-and-Cut solvers employ linear relaxations, leading to exponential growth in auxiliary variables and severe computation limitations.
To overcome these limitations, we propose \underline{\textbf{BIPNN}} (\underline{\textbf{B}}inary \underline{\textbf{I}}nteger \underline{\textbf{P}}rogramming \underline{\textbf{N}}eural \underline{\textbf{N}}etwork), an \textbf{unsupervised} learning framework to solve nonlinear BIP problems via hypergraph neural networks (HyperGNN).
Specifically, \textbf{(\uppercase\expandafter{\romannumeral1})} BIPNN reformulates BIPs-constrained, discrete, and nonlinear ($\mathrm{sin}$, $\mathrm{log}$, $\mathrm{exp}$) optimization problems-into unconstrained, differentiable, and polynomial loss functions.
The reformulation stems from the observation of a precise one-to-one mapping between polynomial BIP objectives and hypergraph structures, enabling the unsupervised training of HyperGNN to optimize BIP problems in an end-to-end manner.
On this basis, \textbf{(\uppercase\expandafter{\romannumeral2})}
we propose a GPU-accelerated and continuous-annealing-enhanced training pipeline for BIPNN.
The pipeline enables BIPNN to optimize large-scale nonlinear terms in BIPs fully in parallel via straightforward gradient descent, thus significantly reducing the training cost while ensuring the generation of discrete, high-quality solutions.
Extensive experiments on synthetic and real-world datasets highlight the superiority of our approach.
\end{abstract}

\section{Introduction}

For decades, binary integer programming (BIP)—a powerful mathematical tool characterized by discrete binary decision variables (0 or 1)—is of critical importance in numerous domains, such as operational optimization~\cite{qiao2021efficient,papalexopoulos2022constrained,wang2024structuredmesh}, quantum computing~\cite{nannicini2022optimal,ajagekar2022hybrid,fan2022hybrid}, computational biology~\cite{llabres2020alignment,zhu2021novel}, materials science and computational chemistry~\cite{gusev2023optimality,stinchfield2024mixed}.
However, BIP is known to be NP-complete~\cite{karp2010reducibility}, making large-scale BIP instances computationally intractable.

Along with AI computing shines in scientific discovery, the potential of neural network-based IP solvers has emerged in recent years.
To address integer linear programming (ILP) problems,
MIP-GNN~\cite{khalil2022mip} leverages graph neural networks (GNN) to improve the performance.
Another GNN\&GBDT-guided framework~\cite{ye2023gnn} for large-scale ILP problems can save up $99\%$ of running time in achieving the same solution quality as SCIP~\cite{MaherMiltenbergerPedrosoRehfeldtSchwarzSerrano2016}, a leading IP solver.
However, these neural network-based ILP solvers lack scalability for nonlinear BIPs.

To handle nonlinearities, state-of-the-art Branch-and-Cut solvers (e.g., SCIP~\cite{achterberg2009scip}) rely on linear relaxation, which introduces a number of auxiliary variables.
Once linearized, these problems are solved using linear programming (LP) solvers (e.g., the Simplex method\footnote{To be precise, the Simplex method is designed to solve linear programming (LP) problems in polynomial time, meaning they belong to the class P~\cite{karmarkar1984new}.}).
Consequently, large-scale nonlinear BIPs often suffer from prohibitive computational costs.
As BIP solvers continue to evolve, linearization remains indispensable for making nonlinearities more tractable for BIP solvers.

\begin{figure}[t]
\centering
\includegraphics[width=1\columnwidth]{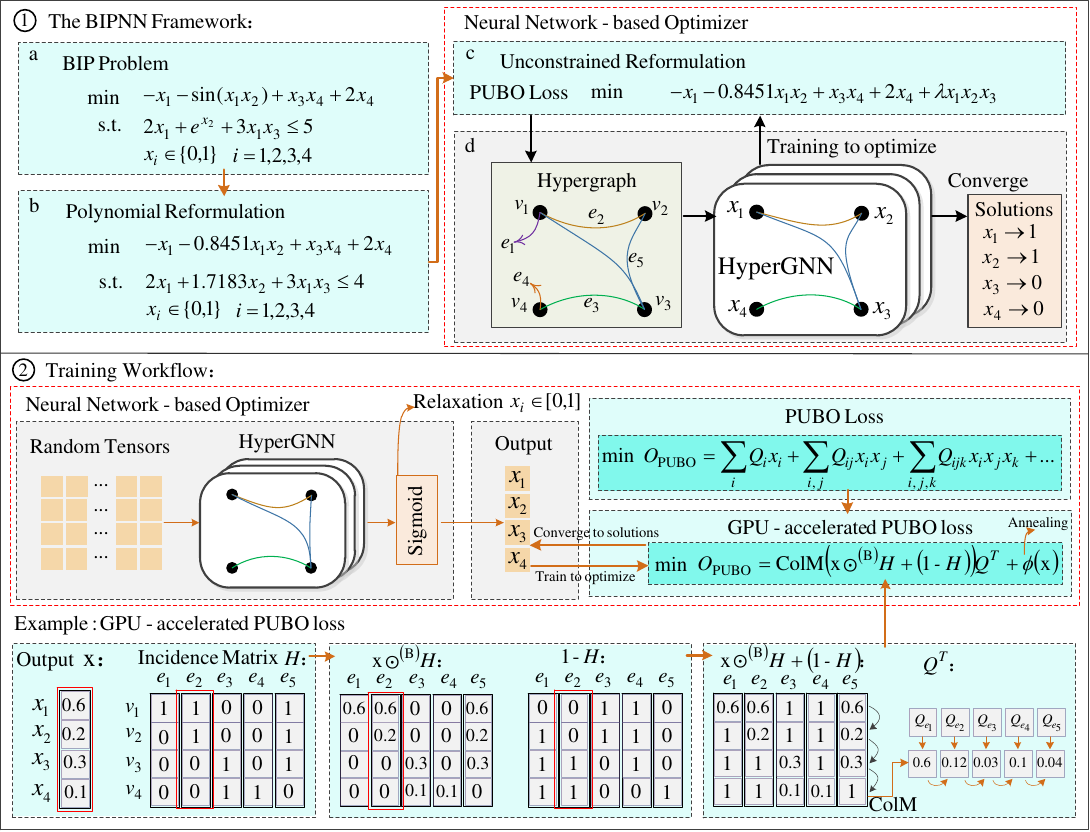}
\caption{The BIPNN framework.}
\label{fig:framework}
\end{figure}
These limitations motivate us to develop a streamlined and general-purpose BIP solver to advance the state of the art.
To profoundly adapt to real-world applications, our work grapples with challenges arising from neural networks' unique characteristics beyond linearization-based methods, as summarized below:

\textbf{Challenge 1}. Meticulously modeling nonlinear terms in BIP objectives and constraints;

\textbf{Challenge 2}. Utilizing GPU's parallel computing capability.

To this end, in this work we propose \underline{\textbf{BIPNN}} (\underline{\textbf{B}}inary \underline{\textbf{I}}nteger \underline{\textbf{P}}rogramming \underline{\textbf{N}}eural \underline{\textbf{N}}etwork), an unsupervised BIP solver that bridges the gap between nonlinear BIP and deep neural networks.
Our overarching idea stems from the observation of one-to-one mapping correspondence between polynomial BIP objectives and hypergraph structures (upper right of Fig.~\ref{fig:framework}). As depicted in Fig.~\ref{fig:framework}, our framework consists of three phases:

1) In the first phase, we employ broadly applicable penalty term method to convert constrained BIP problems into polynomial unconstrained binary optimization (PUBO\footnote{The mathematical formulation PUBO is well-known in quantum computing, for modeling complex optimization problems in a way quantum computers may solve efficiently.}) formalism.
To handle exponential and trigonometric terms, we propose a novel transformation to represent them in the form of polynomials.
These refined polynomial objectives are adaptable to neural network-based solvers when applied as loss functions.

2) In the second phase, we leverage hypergraph neural networks (HyperGNN) to address \textbf{Challenge 1}, capturing high-order correlations between binary decision variables, or in other words the polynomial terms in the refined PUBO objective.
By applying a relaxation strategy to the PUBO objective to generate a differentiable loss function with which we train the HyperGNN in an unsupervised manner.

3) Nevertheless, when we train these HyperGNNs to minimize the PUBO objectives, we encounter severe obstacles of low computational efficiency in these polynomial losses with numerous variables.
In the third phase, leveraging GPUs, we further propose an algorithm to address \textbf{Challenge 2} via matrix operations on the incidence matrices of hypergraphs.

In summary, we contribute:

1) BIPNN, an unsupervised HyperGNN-based solver that allows learning approximate BIP solutions in an end-to-end differentiable way with strong empirical performance.

2) An empirical study of the performance of BIPNN on synthetic and real-world data, demonstrating that unsupervised neural network solvers outperform classic BIP solvers such as SCIP and Tabu in tackling large-scale nonlinear BIP problems.

3) Large-scale nonlinear optimization has long been challenging due to its inherent complexity and scalability issues.
We advance this field by employing several nonlinearity modeling methods for BIP, including the polynomial reformulation and unconstrained reformulation.
These methods provide instructive guidance for unsupervised neural network-based solvers.

\section{Notations and Definitions}
\label{sec:def}
In the following, we will formulate the BIP problem and articulate the definition of hypergraphs.

\textbf{Definition 1} (Formulation of BIP). \textit{Non-linear BIP is an optimization problem where the decision variables $\mathbf{x}=(x_1,x_2,...,x_m)$ are restricted to binary values ($0$ or $1$), and the objective function $O_{\mathrm{BIP}}$ or constraints (or both) are nonlinear. Below is the general formulation.}
\begin{equation}
\begin{gathered}
\mathrm{min} \quad O_{\mathrm{BIP}} = f(\mathbf{x}) \\
\mathrm{s.\ t.} \quad g_k(\mathbf{x}) \leq 0 \quad \mathrm{for\ all}\quad k = 1, 2, \dots, K \\
q_l(\mathbf{x}) =0 \quad \mathrm{for\ all}\quad l = 1, 2, \dots, L \\
x_i \in \{0, 1\} \quad \mathrm{for\ all}\quad i = 1, 2, \dots, n
\end{gathered}
\label{equ:BIP_formulation}
\end{equation}
\textit{where $f(\mathbf{x})$, $g_k(\mathbf{x})$ and $q_l(\mathbf{x})$ are nonlinear functions of the decision variables $\mathbf{x}$.}$\hfill\square$

\textbf{Definition 2} (Hypergraph). \textit{A hypergraph is defined by $G=(V,E)$, where $V=\{v_1,v_2,...,v_{|V|}\}$ stands for a set of vertices and $E=\{e_1,e_2,...,e_{|E|}\}$ denotes a set of hyperedges. Each hyperedge $e_j\in E$ is a subset of $V$. A hypergraph $G$ can be represented by the incidence matrix (Fig.~\ref{fig:framework} at the bottom) $H\in \{0,1\}^{|V|\times |E|}$, where $H_{ij}=1$ if $v_i\in e_j$, or otherwise $H_{ij}=0$.}$\hfill\square$

\section{BIPNN: HyperGNN-based Optimizer for PUBO-formulated BIP}
\label{Sec:BIPNN_phase2}
For easier comprehension of our approach, in this section we first elaborate how to solve an unconstrained, PUBO-formulated BIP problem as depicted in Eq.~\ref{equ:PUBOdef}.
Then, in Sec.~\ref{sec:BIPNN_PUReformulation}, we will show how to transform a general BIP problem with constraints and nonlinear terms into PUBO formalism.

\subsection{Modeling PUBO-formulated BIPs via Hypergraphs}
\label{sec:hypergraph_modeling}
\begin{figure}[t]
\centering
\includegraphics[width=0.9\columnwidth]{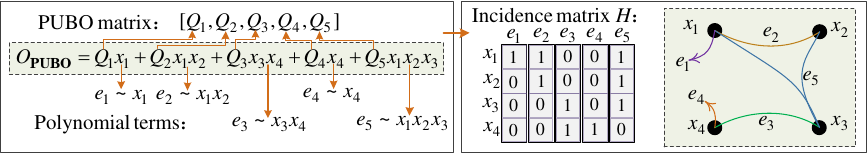}
\caption{Modeling PUBO-formulated BIPs via hypergraphs.}
\label{fig:hypergraph}
\end{figure}
BIPNN employs a HyperGNN-based optimizer (upper right of Fig.~\ref{fig:framework}) to solve PUBO-formulated BIP problems.
Inspired by the binary characteristic of variables, we can reformulate general BIPs as PUBO problems through the polynomial reformulation in Sec.\ref{sec:polynomialReformulation} and unconstrained reformulation in Sec.\ref{sec:UnconstrainedReformulation}.
A PUBO problem is to optimize the cost function:
\begin{equation}
O_{\mathrm{PUBO}}=\sum_{i}{Q_ix_i+\sum_{i,j}{Q_{ij}x_ix_j}}+\sum_{i,j,k}{{Q_{ijk}}{x_ix_jx_k}}+\cdots
\label{equ:PUBOdef}
\end{equation}
where $x_i\in \{0,1\}$ are binary descision variables and the set of all decision variables is denoted by $\mathbf{x}=(x_1,x_2,\cdots,x_m)$.
As shown in Fig.~\ref{fig:hypergraph}, for ease of representation, a PUBO objective $O_{\mathrm{PUBO}}$ with $n$ terms can be decomposed into two components: the PUBO matrix $Q=[Q_1,Q_2,...,Q_n]$, and $n$ linear or polynomial terms such as $x_i$, $x_ix_j$, or $x_ix_jx_k$.
In this way, we discover multi-variable interactions in $O_{\mathrm{PUBO}}$ can be modeled as a hypergraph $G=(V,E)$, where $|E|=n$, and each hyperedge $e\in E$ encodes a single descision variable $x_i$ or a polynomial term such as $x_ix_j$ or $x_ix_jx_k$.

\subsection{Neural Network-based Optimizer}
The training workflow of the neural network-based optimizer is illustrated at the bottom of Fig.~\ref{fig:framework}.

\textbf{HyperGNN Architecture.} Initially, for a PUBO-transformed hypergraph $G=(V,E)$, HyperGNNs take the incidence matrix $H$ of $G$ and a randomly initialized $X^{(0)}\in \mathbb{R}^{m\times d}$ as inputs.
Subsequently, BIPNN applies the sigmoid function to produce the output vector $\mathbf{x}=(x_1,x_2,\cdots,x_m)$, where $x_i\in [0,1]$ are the relaxation of decision variables $x_i\in\{0,1\}$.
The HyperGNN model operates as follows:
\begin{equation}
\mathbf{x}=\mathrm{sigmoid}(\mathrm{HyperGNN}(H,X^{(0)}))
\label{equ:HyperGNN}
\end{equation}
where HyperGNN is a multi-layer hypergraph convolutional network, such as HGNN+~\cite{gao2022hgnn+}, HyperGCN~\cite{yadati2019hypergcn}, or UniGCN~\cite{huang2021unignn}.

\textbf{Training to Optimize.} As an unsupervised learning model, BIPNN relaxes the PUBO objective $O_{\mathrm{PUBO}}$ into a differentiable loss function and trains to optimize it.
Specifically, $O_{\mathrm{PUBO}}$ can be expressed by the output $\mathbf{x}$ and the incidence matrix $H$ as depicted in Fig.~\ref{fig:framework}.
We aim to find the optimal solution $\mathbf{x}^{s}=\mathrm{argmin}O_{\mathrm{PUBO}}(\mathbf{x},H)$.
As training progresses, $x_i\in\mathbf{x}$ will gradually converge to binary solutions.

\textbf{GPU-accelerated Training.} For a large-scale BIP problem, numerous polynomial terms in $O_{\mathrm{PUBO}}$ lead to a high computational cost.
To address this, an intuitive idea is to leverage GPU-supported matrix operations to accelerate training.
However, PUBO problems lack a straightforward matrix formulation.
To this end, we propose GPU-accelerated PUBO objective as follows.
\begin{equation}
O_{\mathrm{PUBO}}=\mathrm{ColM}(\mathbf{x}\odot ^{(\mathrm{B})}H+(1-H))Q^T
\label{equ:GPU_training}
\end{equation}
where $\mathbf{x}$ is the output of HyperGNN, $H$ is the incidence matrix, and $Q=[Q_1,Q_2,...,Q_n]$ is the PUBO matrix.
More concretely, $\mathbf{x}\odot ^{(\mathrm{B})}H$ denotes the element-wise Hadamard product with broadcasting between $m$-dimensional vector $\mathbf{x}$ and matrix $H\in \mathbb{R}^{m\times n}$.
We add $1-H$ on $\mathbf{x}\odot ^{(\mathrm{B})}H$ to fill zero-valued elements with $1$.
Based on this operation, we use the column-wise multiplication denoted by $\mathrm{ColM}$ on the first dimension of the matrix obtained by $\mathbf{x}\odot ^{(\mathrm{B})}H+(1-H)$.
Through the $\mathrm{ColM}$ operation we obtain an $n$-dimensional vector, of which each element represents a polynomial term in $O_{\mathrm{PUBO}}$.
The final loss function is computed by scaling each polynomial term with its respective coefficient $Q_i$.
The detailed explanation is illustrated in Fig.~\ref{fig:framework}.

\textbf{Time Complexity Analysis.} For $\mathbf{x}\in \mathbb{R}^{m}$, $Q\in \mathbb{R}^{1\times n}$, and $H\in \mathbb{R}^{m\times n}$, the time complexity of Eq.~\ref{equ:GPU_training} is $O(m\times n)$.
For GPU-accelerated training, element-wise operations such as Hadamard product are fully parallelizable.
Column-wise product over $m$ leads to time complexity $O(\mathrm{log}m)$. Thus, the theoretical best GPU time complexity is $O(\mathrm{log}m)$.
Utilizing $T$ cores, the realistic GPU time complexity is $O(\frac{m\times n}{T})$.

\textbf{Annealing Strategy.} To achieve unsupervised learning, BIPNN relaxes PUBO problems into continuous space. The differentiable relaxation of discrete decision variables sometimes leads to continuous solutions $x_i\in [0,1]$.
To address this, we employ the continuous relaxation annealing (CRA)~\cite{ichikawa2024controlling} method.
Specifically, BIPNN uses the following loss function: $O_{\mathrm{PUBO}}=\mathrm{ColM}(\mathbf{x}\odot ^{(\mathrm{B})}H+(1-H))Q^T+\phi(\mathbf{x})$, where $\phi(\mathbf{x})=\gamma \sum_{i=1}^{n}(1-(2x_i-1)^{\alpha})$ is the penalty term, $\gamma$ controls the penalty strength and $\alpha$ is an even integer.
We initialize $\gamma<0$ and gradually increase it to a positive value as training progresses.
The annealing strategy enhances the performance of BIPNN in three aspects, \textbf{(\romannumeral1)} In the high-temperature phase ($\gamma<0$), it smooths the HyperGNN, preventing it from getting trapped in local optima; \textbf{(\romannumeral2)} In the low-temperature phase ($\gamma>0$), it enforces the discreteness of solutions; \textbf{(\romannumeral3)} It effectively accelerates the training process.

\section{BIPNN: Polynomial \& Unconstrained Reformulation of BIP}
\label{sec:BIPNN_PUReformulation}
In this section, we explain how to reformulate nonlinear BIPs as unconstrained and polynomial optimization problems, which are compatible with our neural network-based optimizer.

\subsection{Polynomial Reformulation of BIP}
\label{sec:polynomialReformulation}
Our approach is inspired by the observation that for any binary variable, a nonlinear term such as $e^x$ can be exactly fitted by a polynomial equivalent $h(x)=ax+b$, such that $h(x)=e^x$ for $x\in \{0,1\}$. That is, $h(x)=(e-1)x+1$, where $h(0)=1$ and $h(1)=e$.
To handle univariate nonlinearities, including trigonometric, logarithmic, and exponential terms (e.g., $\mathrm{sin}x$, $\mathrm{log}x$, and $e^x$), we have the following transformation: $h(x)=(h(1)-h(0))x+h(0)$.
For multivariate terms such as $e^{x_ix_j}$ and $sin(x_ix_j)$, where $x_ix_j\in\{0,1\}$, we can perform the transformation as follows: $h(\prod_{i \in S} x_i)=(h(1)-h(0))\prod_{i \in S} x_i+h(0)$.

BIPNN employs a more general method to handle more intricate multivariate nonlinear terms (such as $\mathrm{sin}(x_i+x_j)$).
For a set of binary decision variables $x_1,x_2,...,x_n$, a non-linear function $h(x_1,x_2,...,x_n)$ can be transformed into the polynomial forms as follows.
\begin{equation}
h(x_1,x_2,...,x_m)=\sum_{S\subseteq \{1,2,...,m\}}{c_S\prod_{i\in S}{x_i}}
\label{equ:BNPF_1}
\end{equation}
By setting up a system of equations based on all possible combinations of $x_1,x_2,...,x_m$, we can determine the coefficients $c_S$ to precisely fit $h(x_1,x_2,...,x_m)$ by leveraging simple inclusion-exclusion principle (refer to Appendix A) as below.
\begin{equation}
c_S=\sum_{T\subseteq S}{(-1)^{|S|-|T|}f(T)}
\label{equ:BNPF_2}
\end{equation}
where $f(T)$ represents the function value when the variables in the subset $T$ are $1$ and the others are $0$.
For each subset $S$, it needs to calculate $2^{|S|}$ values of $f(T)$.$\hfill\square$

As an example, we have $\mathrm{sin}(x_1+x_2)=0.8415x_1+0.8415x_2-0.7737x_1x_2$. A toy example of $\mathrm{sin}(x_1+x_2+x_3)$ is illustrated in Appendix A.
To be noticed, polynomial reformulation of all nonlinear terms in a BIP objective is not necessary.
If the transformation becomes overly complex, we may opt to retain the original nonlinear term and directly incorporate it as part of the loss function of HyperGNN.

\subsection{Unconstrained Reformulation of BIP}
\label{sec:UnconstrainedReformulation}
We propose a novel penalty method to transform the constrained BIP problem into an unconstrained form.
In penalty methods~\cite{nocedal1999numerical,glover2022quantum}, unconstrained reformulation is achieved by adding "penalty terms" to the objective function that penalize violations of constraints.
A well-constructed penalty term must be designed such that it equals $0$ if and only if the constraint is satisfied, and takes a positive value otherwise.
Specifically, given a BIP problem in Eq.~\ref{equ:BIP_formulation}, for inequality constraints $g_k(\mathbf{x})\leq 0$, we have penalty terms $P_k(\mathbf{x})=\lambda_k \cdot \left( \max\left(0, g_k(\mathbf{x})\right) \right)^2$,
for equality constraints $q_l(\mathbf{x})=0$, we have penalty terms $Q_l(\mathbf{x})=\mu_l \cdot \left( q_l(\mathbf{x}) \right)^2$, where \( \lambda_k, \mu_l \) are sufficiently large penalty coefficients. By combining all terms into a single objective function, we have an unconstrained BIP objective:
\begin{equation}
\text{min} \quad O_{\mathrm{BIP}}=f(\mathbf{x}) + \sum_{k=1}^{K}{\lambda_k \cdot \left( \max\left(0, g_k(\mathbf{x})\right) \right)^2} + \sum_{l=1}^{L}{\mu_l \cdot \left( q_l(\mathbf{x}) \right)^2}
\label{equ:BIPPenalty_1}
\end{equation}
As part of the loss function of BIPNN, $O_{\mathrm{BIP}}$ must be differentiable to enable gradient-based optimization.
However, $\max\left(0, g_k(\mathbf{x})\right)$ is not a continuously differentiable function, thus finding an appropriate penalty term is crucial.
We propose two methods to address this issue:

1) ${\mathrm{\textbf{ReLU}}}$\textbf{-based Penalty}. We can use $\mathrm{ReLU}{(g_k(\mathbf{x}))^2}=({\mathrm{max}(0,g_k(\mathbf{x}))})^2$ to handle constraints. This is a general method for a large number of variables $x_i$ in a constraint $g_k(\mathbf{x})$.

2) \textbf{Polynomial Penalty}. In the following, we present an algorithm to construct polynomial penalty terms with $2^\Delta$ time complexity for $g_k(\mathbf{x})$, where $\Delta$ is the number of variables in constraint $g_k(\mathbf{x})$.

For binary variables, do there exist polynomial penalty terms that correspond to BIP constraints? To answer this question, we have the following discussion.
%For example, binary variables are denoted by \( x_i \in \{0,1\} \), the penalty term of constraint $x_1x_2x_3\leq 0$ derived from $P_k(\mathbf{x})=\lambda_k \cdot \left( \max\left(0, g_k(\mathbf{x})\right) \right)^2$ is $\lambda x_1x_2x_3$.
%However, the penalty term $\lambda x_1x_2$ for constraint $x_1+2x_2-2\leq 0$, cannot be trivially derived.
For $x_1+2x_2-2\leq 0$, we observe that the violating subset $\{x_1=1,x_2=1\}$ corresponds to polynomial penalty term $\lambda(x_1x_2)$. For another constraint $x_1+3x_2-2\leq 0$, the violating subsets $\{x_1=0,x_2=1\}$ and $\{x_1=1,x_2=1\}$ correspond to the polynomial penalty term $\lambda(x_2+x_1x_2)$ or $\lambda x_2$.
Through an in-depth analysis, we propose a novel method to transform nonlinear BIP constraints into polynomial penalty terms.
To handle an inequality constraint $g(\mathbf{x})\leq 0$ for the BIP problem in Eq.~\ref{equ:BIP_formulation}, our method consists of three steps (to see a toy example, refer to Appendix B):

\textbf{(\romannumeral1)} Initially, we express the constraint $g(\mathbf{x})\leq 0$ as a boolean indicator function: $\psi(\mathbf{x}) = \begin{cases} 
1 & \text{if } g(\mathbf{x}) > 0 \ (\text{violation}) \\
0 & \text{otherwise} \ (\text{feasible})
\end{cases}$, then define minimal violation subsets $\mathcal{V}$ as the smallest variable combinations causing constraint violations:
\begin{equation}
\mathcal{V} = \left\{ S \subseteq \{1,...,n\} \, \bigg| \, \psi(\mathbf{x}) = 1 \text{ when } x_i = 1 \ \forall i \in S \text{ and } x_j = 0 \ \forall j \notin S \right\}
\label{equ:Penalty_3}
\end{equation}
each $S\in \mathcal{V}$ cannot be reduced further without eliminating the violation.

\textbf{(\romannumeral2)} Generate a penalty term for each minimal violation subset $S\in \mathcal{V}$:
\begin{equation}
P(\mathbf{x}) = \lambda \sum_{S \in \mathcal{V}} \prod_{i \in S} x_i
\label{equ:Penalty_4}
\end{equation}
where $\lambda$ is the penalty coefficient.

\textbf{(\romannumeral3)} Combine each term into the BIP objective function:
\begin{equation}
\text{min} \quad O_{\mathrm{BIP}} = f(\mathbf{x}) + P(\mathbf{x})
\label{equ:Penalty_5}
\end{equation}
In the worst case, when an enumeration method is used in step \textbf{(\romannumeral1)}, it requires calculating $2^\Delta$ subsets, where $\Delta$ is the number of variables in constraint $g(\mathbf{x})$.
Nevertheless, in most real-world problems (e.g. max-cut, and maximal independent set or MIS) involving graphs, the variables associated with each constraint often exhibit locality.$\hfill\square$

\begin{figure*}[t]
\centering
\includegraphics[width=1\columnwidth]{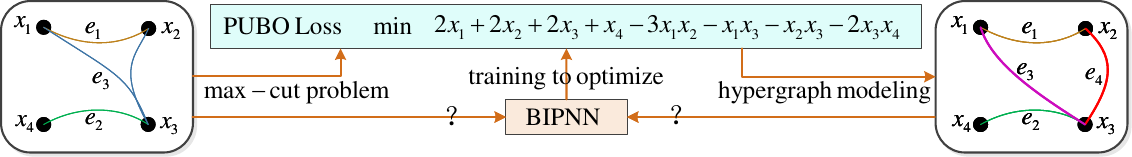}
\caption{To solve the hypergraph max-cut problem, BIPNN generates a new hypergraph structure. However, both of these hypergraphs can be utilized for training the HyperGNN model.}
\label{fig:hypergraph_generate}
\end{figure*}
The polynomial penalty method facilitates to incorporate penalty terms to PUBO objectives and use GPU-accelerated training pipeline to solve BIPs.
As far as we know, only a few number of constraint/penalty pairs~\cite{glover2022quantum} associated have been identified in existing literature.
Our work significantly expands the potential application domains of the penalty method.
%Our work provides a novel unconstrained reformulation solution to address this issue, thereby significantly expanding the potential application domains of neural network or quantum annealing-based BIP solvers.

\section{Discussion}
\label{sec:discussion}
\textbf{Feasible Solutions.} Firstly, a PUBO problem always has feasible solutions.
The feasible set is the entire space of binary variable combinations, since there are no constraints to exclude any combination. Every possible binary assignment $x_i\in\{0,1\}$ is inherently feasible.
Secondly, the feasibility of a nonlinear BIP problem depends on the constraint compatibility—whether there exists at least one binary variable assignment $\mathbf{x}\in\{0,1\}^m$ that satisfies all nonlinear constraints simultaneously.
In BIPNN, we determine the existence of feasible solutions through \textbf{(\romannumeral1)} Training-phase feasibility check: if all penalty terms (e.g., constraint violations) converge to zero during training, feasible solutions exist; otherwise, the problem is infeasible.
\textbf{(\romannumeral2)} Post-training verification: we sample candidate solutions from the trained model and explicitly verify whether they satisfy all constraints.

\textbf{The Effectiveness of BIPNN's Hypergraph Generation Mechanism.} As depicted in Fig.~\ref{fig:hypergraph_generate}, when BIPNN is applied to solve combinatorial optimization (CO) problems on hypergraphs, it generates an alternative hypergraph structure.
However, both of the hypergraphs can be used as the input of BIPNN.
A critical question arises: which type of hypergraph structure achieves better performance when applied to HyperGNN?
The main difference between these two hypergraphs is that the hypergraph generated by BIPNN breaks down the original hypergraph's high-order hyperedges into numerous low-order ones.
We argue that BIPNN training with the original hypergraph structure is more computationally efficiency, while BIPNN-generated hypergraph structure leads to more optimal solutions.
In Sec.~\ref{sec:expHypergraphGeneration}, we will empirically compare the solution quality of both methods.
\section{Experimental Results}
\label{sec:experiment}
\begin{figure}[t]
\centering
\begin{minipage}[b]{0.32\textwidth}
    \centering
    \includegraphics[width=\linewidth]{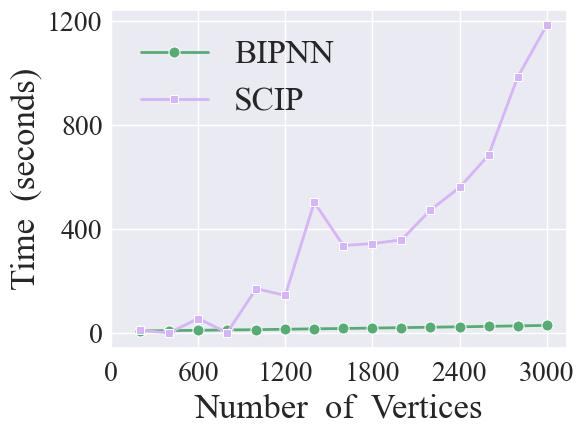}
    \subcaption{SCIP, $d=4$.}
    \label{fig:SCIPtimek4}
\end{minipage}
\hfill
\begin{minipage}[b]{0.32\textwidth}
    \includegraphics[width=\linewidth]{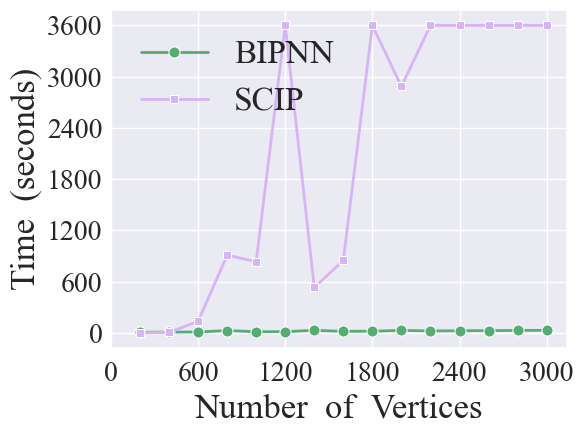}
    \subcaption{SCIP, $d=6$.}
    \label{fig:SCIPtimek6}
\end{minipage}
\hfill
\begin{minipage}[b]{0.32\textwidth}
    \includegraphics[width=\linewidth]{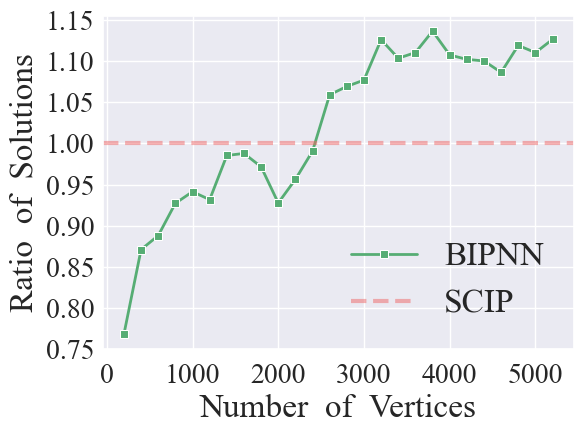}
    \subcaption{SCIP, $d=4$.}
    \label{fig:SCIPsolutionratiok4}
\end{minipage}
\hfill
\begin{minipage}[b]{0.32\textwidth}
    \centering
    \includegraphics[width=\linewidth]{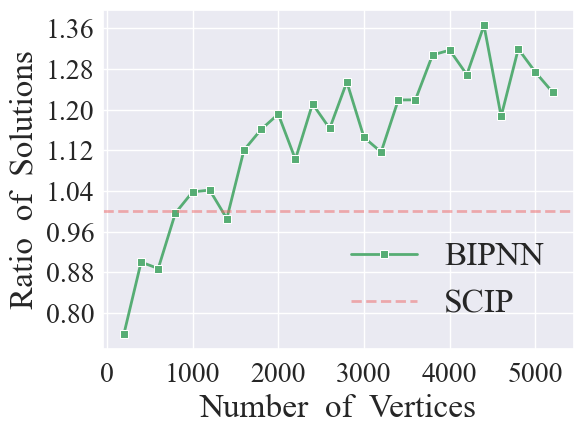}
    \subcaption{SCIP, $d=6$.}
    \label{fig:SCIPsolutionratiok6}
\end{minipage}
\hfill
\begin{minipage}[b]{0.32\textwidth}
    \includegraphics[width=\linewidth]{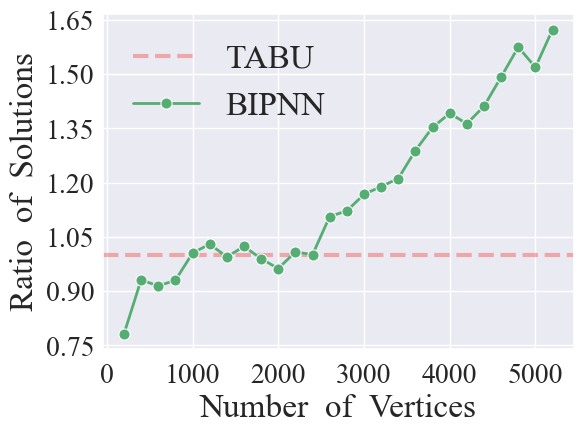}
    \subcaption{Tabu, $d=4$.}
    \label{fig:Tabusolutionratiok4}
\end{minipage}
\hfill
\begin{minipage}[b]{0.32\textwidth}
    \includegraphics[width=\linewidth]{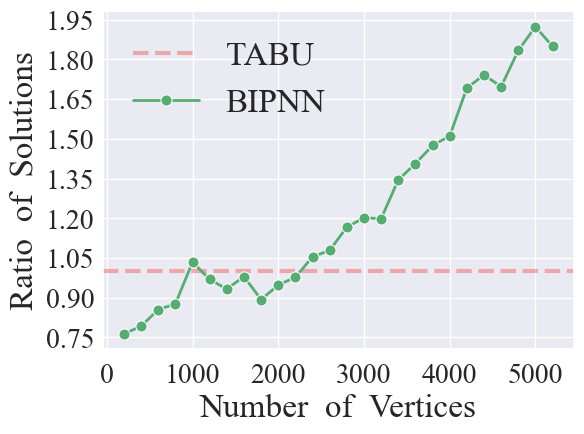}
    \subcaption{Tabu, $d=6$.}
    \label{fig:PyQUBOsolutionratiok4}
\end{minipage}
\caption{Comparison of BIPNN and existing BIP solvers. $d$ is the degree of polynomial terms in BIP objective functions. (a)(b) show the solving time required for BIPNN and SCIP to obtain the same solution. (c)(d) show the ratio of the solutions of BIPNN to SCIP; (e)(f) illustrate the ratio of the solutions of BIPNN to Tabu; Runtime is restricted to 1 hour.}
\end{figure}
In this section, we describe our empirical experiments on BIPNN and baseline optimization tools.
%We release the implementation of BIPNN at this URL\footnote{TODO BIPNN github}.
%The project provides utility programs for unconstrained and polynomial reformulation of general BIP problems, hence our focus lies in assessing model quality of BIPNN in solving PUBO problems.

\textbf{Benchmarks}.
To evaluate BIPNN on BIP problems with diverse scales, the datasets are generated using DHG library\footnote{https://deephypergraph.readthedocs.io/en/latest/index.html}.
To evaluate the quality of solutions and computational efficiency of BIPNN, datasets of varying scales are generated in three steps: Initially, DHG library is applied to generate hypergraph structures (where $|E|=2|V|$). Subsequently, a random coefficient is assigned to each hyperedge (representing a polynomial term) to generate PUBO objective functions.
Thereafter, several constraints (penalty terms) were randomly incorporated into the PUBO objectives.
To demonstrate the effectiveness of BIPNN on real-world settings, we also conduct experiments on the hypergraph max-cut problem (refer to Appendix~\ref{sec:hypergraphmaxcut}), a well-known BIP problem benchmark.
Moreover, we conduct experiments on publicly-available hypergraph datasets (refer to Appendix~\ref{app:datasets}).
%including Primary, Cora, and Pubmed.

\textbf{Baseline Methods.} In our experiments, the baseline methods include optimization techniques and tools such as SCIP~\cite{MaherMiltenbergerPedrosoRehfeldtSchwarzSerrano2016}, Tabu search~\cite{glover1998tabu}.

\textbf{Implementation Details}. Experiments are conducted on an Intel Core i9-12900K CPU with 24 cores, and an NVIDIA GeForce RTX 3090 GPU with 24 G of memory.
We adopt two-layer HGNN+~\cite{gao2022hgnn+} as the HyperGNN model for the experiments.

\subsection{Comparison with Linearization-based BIP Solvers}
\textbf{SCIP.} SCIP is an exact solver based on the branch-and-cut algorithm.
Theoretically, given sufficient time and computational resources, SCIP guarantees an exact solution.
However, for large-scale problems, due to time constraints, SCIP may terminate prematurely and return the approximate solution.
To conduct the experiment, we generate a specific BIP instance for each size of variables.
Specifically, for a BIPNN-generated hypergraph, the number of vertices (variables) $|V|$ ranges from $200$ to $3000$.
The degrees of vertices are set to $4$ (Fig.~\ref{fig:SCIPtimek4}) and $6$ (Fig.~\ref{fig:SCIPtimek6}) respectively.

Fig.~\ref{fig:SCIPtimek4} and Fig.~\ref{fig:SCIPtimek6} show the comparison of the solving time for BIPNN and SCIP.
We evaluate the solving time taken by BIPNN to obtain the best approximate solution and the time required by SCIP to find the same solution.
Experimental results demonstrate that the solving time of BIPNN grows linearly and slowly with increasing problem size, while SCIP's solving time exhibits exponential growth.
This trend becomes more pronounced when the degree of polynomial terms is $6$.

Moreover, we impose a $1$-hour time limit and evaluate the solution quality of BIPNN and SCIP across varying scales of BIP instances.
Fig.~\ref{fig:SCIPsolutionratiok4} and Fig.~\ref{fig:SCIPsolutionratiok6} show the comparative  ratio of solutions obtained by BIPNN and SCIP.
Specifically, the comparative ratio is defined as $\frac{O^s_{\mathrm{BIPNN}}}{O^s_{\mathrm{SCIP}}}$, where $O^s_{\mathrm{BIPNN}}$ and $O^s_{\mathrm{SCIP}}$ are the solutions obtained by BIPNN and SCIP.
Experimental results demonstrate that BIPNN starts outperforming SCIP when the number of variables exceeds $2,500$ when $d=4$.
As the problem size increases, BIPNN's solutions increasingly outperform SCIP's solutions.
For $d=6$, BIPNN outperforms SCIP when the number of vertices exceeds $1,000$.

\textbf{Tabu Search.} Tabu search is a heuristic method that typically provides approximate solutions.
We also impose a $1$-hour time limit and evaluate the difference in solution quality for Tabu when the degrees of polynomial terms are set to $4$ and $6$.
The number of vertices (variables) $|V|$ in the hypergraph generated by BIPNN ranges from $200$ to $5,000$.
Experimental results are depicted in Fig.~\ref{fig:Tabusolutionratiok4} ($d=4$) and Fig.~\ref{fig:PyQUBOsolutionratiok4} ($d=6$).
As shown in the figures, BIPNN achieves the performance comparable to Tabu when the number of variables exceeds $1,000$. When the number of variables exceeds $2,500$, BIPNN significantly outperforms Tabu as the variable count increases further.
%Fig.~\ref{fig:PyQUBOsolutionratiok4} shows that although PyQUBO can convert PUBO problems into quadratic-cost problems for solving, this approach—like linearization-based methods—introduces excessive auxiliary variables, significantly degrading solution quality. As a result, BIPNN achieves far superior solution quality compared to PyQUBO.
%比较cora，pubmed，cooking200三个数据的MIS和max-cut（不知道能不能做，试试看）的解和运行时间，画成表
%real-world data 用图3的两种hypergraph结构分别测试一遍
\subsection{Comparison on Real-world Datasets}
\begin{table*}[t]
\caption{The solutions of graph/hypergraph max-cut problems ($1$-hour time limit).}
\small
\centering
\begin{tabular}{cccccccccc}
\toprule
Method&
BAT&
EAT&
UAT&
DBLP&
CiteSeer&
Primary&
High&
Cora\\
\midrule
SCIP&
\underline{655}&
3,849&
7,899&
\underline{2,869}&
\underline{3,960}&
7,603&
4,599&
1,215\\
Tabu&
652&
3,972&
8,402&
2,710&
3,717&
8,500&
5,160&
1,360\\
BIPNN&
651&
\textbf{3,978}&
\textbf{8,407}&
2,801&
3,852&
\textbf{8,509}&
\textbf{5,216}&
\textbf{1,384}\\
\bottomrule
\end{tabular}
%}
\label{tab:realworlddata}
\end{table*}
\begin{figure}[t]
\begin{minipage}[b]{0.24\textwidth}
    \centering
    \includegraphics[width=\linewidth]{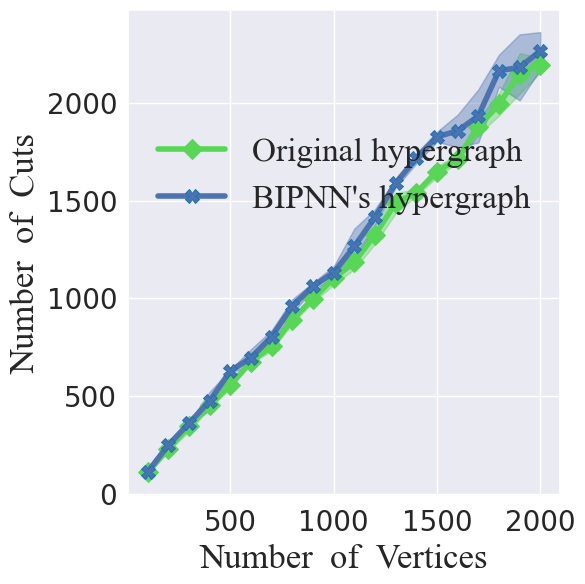}
    \subcaption{$d=4$.}
    \label{fig:HGM1}
\end{minipage}
\hfill
\begin{minipage}[b]{0.24\textwidth}
    \centering
    \includegraphics[width=\linewidth]{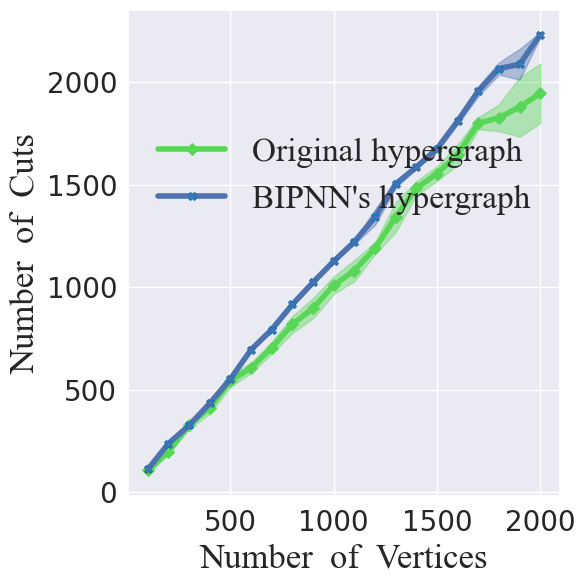}
    \subcaption{$d=6$.}
    \label{fig:HGM2}
\end{minipage}
\hfill
\begin{minipage}[b]{0.24\textwidth}
    \includegraphics[width=\linewidth]{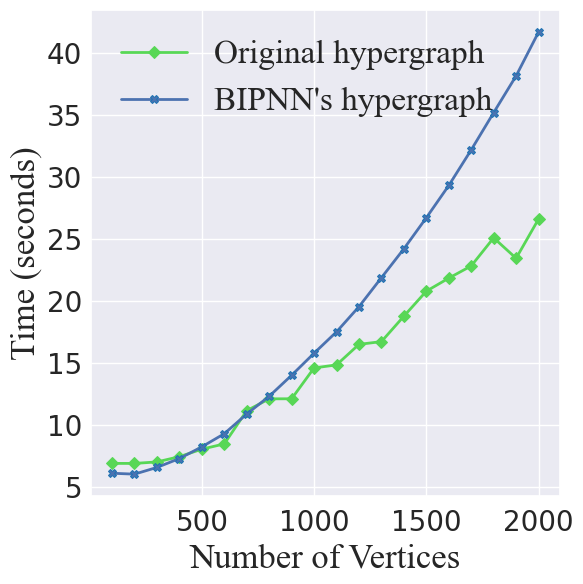}
    \subcaption{$d=4$.}
    \label{fig:HGM3}
\end{minipage}
\hfill
\begin{minipage}[b]{0.24\textwidth}
    \includegraphics[width=\linewidth]{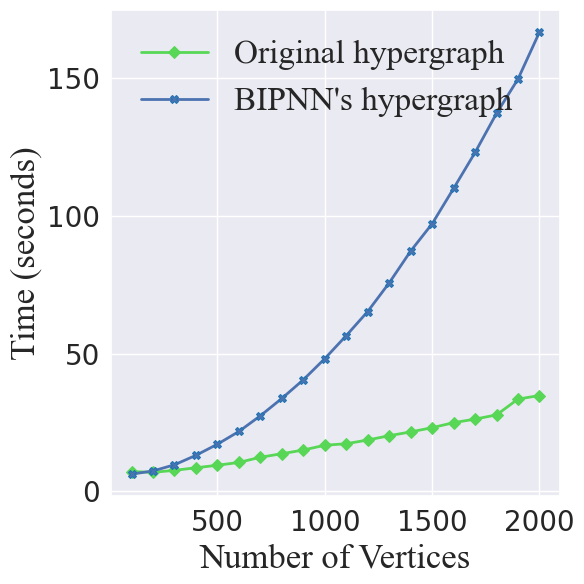}
    \subcaption{$d=6$.}
    \label{fig:HGM4}
\end{minipage}
\caption{Comparison of the quality of solutions and time efficiency of BIPNN when it applys its generated hypergraph structure or the original hypergraph structure to solve hypergraph max-cut problems. $d$ is the degree of polynomial terms in BIP objective functions. (a)(b) show the numbers of cuts; (c)(d) show the solving time.}
\label{fig:hypergraphgenerationmechanism}
\end{figure}
We compare our method against baseline methods on real-world graph and hypergraph datasets, including BAT, EAT, UAT, DBLP, CiteSeer, Primary, High, and Cora (refer to Appendix~\ref{app:datasets}).
Graph datasets include BAT, EAT, UAT, DBLP, and CiteSeer.
Hypergraph datasets include Primary, High, and Cora.
Graph and hypergraph max-cut problems are selected as the BIP problem benchmarks.
We impose $1$ hour time limit and evaluate the number of cuts obtained by BIPNN, SCIP, and Tabu.
%We also evaluate the performance of BIPNN with both its generated hypargraph structures and the original graph/hypergraph structures.
As depicted in Tab.~\ref{tab:realworlddata}, SCIP achieved the best performance on three graph datasets, while BIPNN achieved the best performance on two graph datasets and all three hypergraph datasets.
In summary, compared to the graph max-cut problem, due to higher degree of polynomial terms in the objective function of the hypergraph max-cut problem, BIPNN tends to achieve better performance on hypergraph datasets.
\subsection{Comparative Analysis on Hypergraph Generation Mechanism}
\label{sec:expHypergraphGeneration}
In Sec.~\ref{sec:discussion} and Fig.~\ref{fig:hypergraph_generate}, we propose to evaluate the effectiveness of BIPNN's hypergraph generation mechanism by comparing the effects of its generated hypergraph structures against the original hypergraph structures in a hypergraph CO problem.
In this section, we select hypergraph max-cut as benchmark and conduct experiments to evaluate the performance of BIPNN under both of the hypergraph structures.
Experimental results are depicted in Fig.~\ref{fig:hypergraphgenerationmechanism}.
The number of variables ranges from $100$ to $2000$.
The degrees of polynomial terms $d$ are set to $d=4$ and $d=6$ respectively.
We perform 10 tests each time and record the average value of the cut numbers.
As illustrated in Fig.~\ref{fig:HGM1} and Fig.~\ref{fig:HGM2}, the hypergraph structure generated by BIPNN can identify more cuts in comparison.
However, as depicted in Fig.~\ref{fig:HGM3} and Fig.~\ref{fig:HGM4}, when the parameter $d$ is larger, the number of hyperedges (polynomial terms in PUBO objectives) in the hypergraph structure generated by BIPNN increases sharply, leading to significantly higher computational costs.
The results align with the theoretical analysis we presented in Sec.~\ref{sec:discussion}.
\begin{wrapfigure}[14]{r}{4.5cm}
  \includegraphics[width=4.3cm]{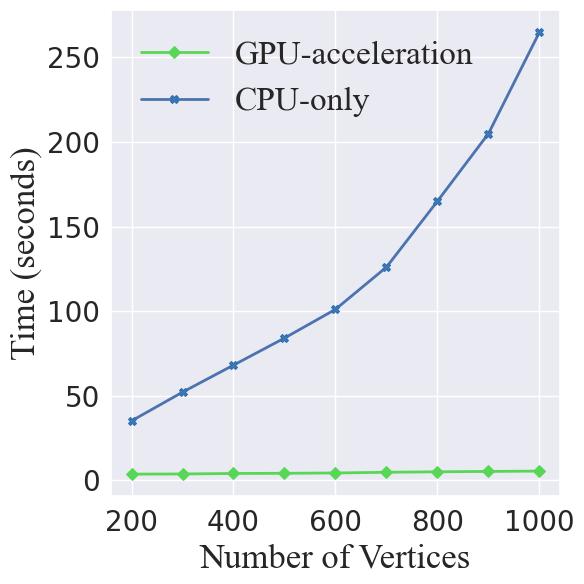}
  \caption{Comparison of the training time for BIPNN with or without GPU accelerated algorithm for PUBO losses.}
  \label{fig:AblationGPU}
\end{wrapfigure}
\subsection{Ablation Study}
\textbf{GPU Acceleration.}
The superior time efficiency of BIPNN is primarily attributed to the GPU-accelerated algorithm employed in computing large-scale PUBO loss functions.
Fig.~\ref{fig:AblationGPU} shows a comparison of the training times for BIPNN with or without the GPU-accelerated algorithm.
We evaluate the training time of BIPNN on the hypergraph max-cut problem.
The number of variables ranges from $200$ to $1000$.
The degree of polynomial terms is set to $4$.
We train BIPNN for a fixed number of 1000 epochs.
As Fig.~\ref{fig:AblationGPU} illustrates, when GPU acceleration is applied to compute the PUBO loss function, the training time does not exhibit significant growth with an increasing number of variables. In contrast, without GPU acceleration, the training time increases rapidly as the number of variables rises.

\begin{wrapfigure}[13]{r}{4.5cm}
  \includegraphics[width=4.3cm]{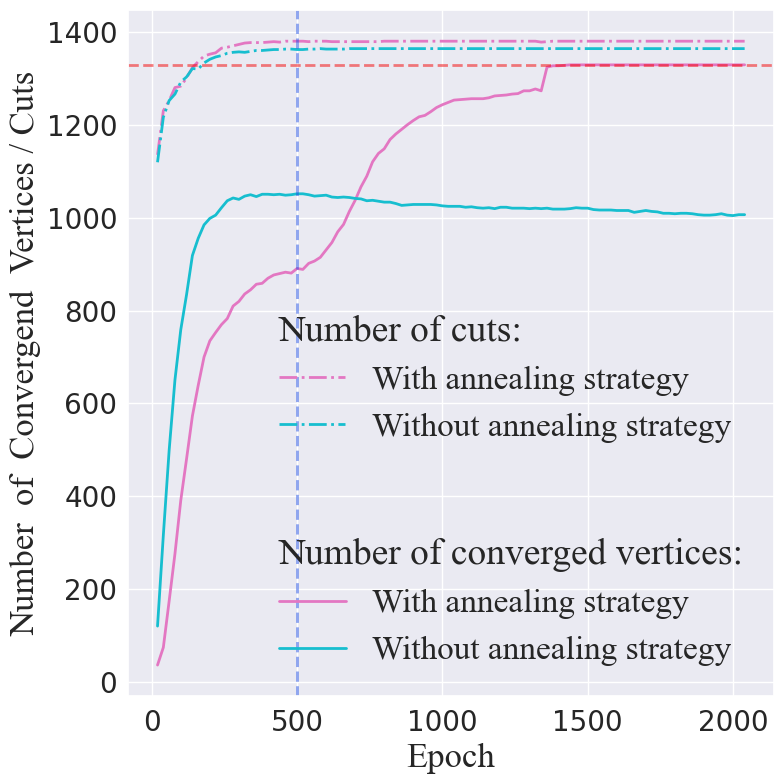}
  \caption{Quality and discreteness of solutions with or without the annealing strategy.}
  \label{fig:AblationAnnealing}
\end{wrapfigure}
\textbf{Annealing Strategy.}
We validate the effectiveness of the annealing strategy of BIPNN on the hypergraph max-cut problem.
The experiments are conducted on Cora with $1,330$ vertices.
The metrics include the number of cuts and discreteness of variables.
The penalty strength $\gamma$ is set to $-2.5$ initially and its value is gradually increased during training.
The value of $\gamma$ reaches $0$ after $500$ epochs and continued to increase thereafter.
As illustrated in Fig.~\ref{fig:AblationAnnealing}, the annealing strategy ensures BIPNN to get better solutions while guaranteeing all variables to converge to discrete values.
It demonstrates that negative $\gamma$ values enable BIPNN to escape local optima, thereby discovering better solutions.
Moreover, when $\gamma$ is set to positive values, it facilitates the convergence of variables toward discrete values.

\section{Conclusion}
\label{sec:conclusion}
This work proposes BIPNN, a novel neural network solver for nonlinear BIP problems.
It reformulates nonlinear BIPs into PUBO cost functions, which correspond to hypergraph structures.
%BIPNN employs HyperGNNs to capture polynomial 
On this basis, these PUBO cost functions are used as loss functions for HyperGNNs, enabling the model to solve BIPs in an unsupervised training manner.
Compared with existing BIP solvers (e.g., SCIP) that rely on linearization, BIPNN reduces the training cost by optimizing nonlinear BIPs via straightforward gradient descent.
Empirical results demonstrate that BIPNN achieves state-of-the-art performance in learning approximate solutions for large-scale BIP problems.

\newpage
\bibliographystyle{unsrt}
\bibliography{citation}

%%%%%%%%%%%%%%%%%%%%%%%%%%%%%%%%%%%%%%%%%%%%%%%%%%%%%%%%%%%%%%%%%%%%%%%%%%%%%%%
%%%%%%%%%%%%%%%%%%%%%%%%%%%%%%%%%%%%%%%%%%%%%%%%%%%%%%%%%%%%%%%%%%%%%%%%%%%%%%%
% APPENDIX
%%%%%%%%%%%%%%%%%%%%%%%%%%%%%%%%%%%%%%%%%%%%%%%%%%%%%%%%%%%%%%%%%%%%%%%%%%%%%%%
%%%%%%%%%%%%%%%%%%%%%%%%%%%%%%%%%%%%%%%%%%%%%%%%%%%%%%%%%%%%%%%%%%%%%%%%%%%%%%%
\newpage
\appendix
\onecolumn

\section{A toy example of the polynomial reformulation of BIP (Sec.~\ref{sec:BIPNN_PUReformulation}).}

For $\sin(x_1 + x_2 + x_3)$, where $x_1, x_2, x_3 \in \{0, 1\}$, we can construct a polynomial to precisely fit the function, such that it matches $\sin(x_1 + x_2 + x_3)$ for all combinations of $x_1, x_2, x_3 \in \{0, 1\}$. For multiple binary variables, the polynomial can be generalized as:
\begin{equation}
P(x_1, x_2, x_3) = a_1 x_1 + a_2 x_2 + a_3 x_3 + b_{12} x_1 x_2 + b_{13} x_1 x_3 + b_{23} x_2 x_3 + c x_1 x_2 x_3 + d
\end{equation}

Based on all possible combinations of $x_1, x_2, x_3$, we can set up the following equations:

1) When $x_1 = 0, x_2 = 0, x_3 = 0$: $P(0, 0, 0) = d = \sin(0) = 0$. Thus, $d = 0$.

2) When $x_1 = 0, x_2 = 0, x_3 = 1$: $P(0, 0, 1) = a_3 = \sin(1) \approx 0.8415$. Thus, $a_3 = 0.8415$.

3) When $x_1 = 0, x_2 = 1, x_3 = 0$: $P(0, 1, 0) = a_2 = \sin(1) \approx 0.8415$. Thus, $a_2 = 0.8415$.

4) When $x_1 = 1, x_2 = 0, x_3 = 0$: $P(1, 0, 0) = a_1 = \sin(1) \approx 0.8415$. Thus, $a_1 = 0.8415$.

5) When $x_1 = 0, x_2 = 1, x_3 = 1$: $P(0, 1, 1) = a_2 + a_3 + b_{23} = \sin(2) \approx 0.9093$.

Substituting $a_2 = 0.8415$ and $a_3=0.8415$: $b_{23}=-0.7737$.

6) When $x_1 = 1, x_2 = 0, x_3 = 1$: $P(1, 0, 1) = a_1 + a_3 + b_{13} = \sin(2) \approx 0.9093$

Substituting $a_1 = 0.8415$ and $a_3 = 0.8415$: $b_{13}=-0.7737$.

7) When $x_1 = 1, x_2 = 1, x_3 = 0$: $P(1, 1, 0) = a_1 + a_2 + b_{12} = \sin(2) \approx 0.9093$

Substituting $a_1 = 0.8415$ and $a_2 = 0.8415$: $b_{12} = -0.7737$

8) When $x_1 = 1, x_2 = 1, x_3 = 1$: $P(1, 1, 1) = a_1 + a_2 + a_3 + b_{12} + b_{13} + b_{23} + c = \sin(3) \approx 0.1411$.

Substituting known values: $c = -0.0623$.

Based on the above calculations, the polynomial is:
\begin{equation}
P(x_1, x_2, x_3) = 0.8415 (x_1 + x_2 + x_3) - 0.7737 (x_1 x_2 + x_1 x_3 + x_2 x_3) - 0.0623 x_1 x_2 x_3
\end{equation}

\section{A toy example of the unconstrained reformulation of BIP (Sec.~\ref{sec:BIPNN_PUReformulation}).}
For a nonlinear constraint with exponential term $g(\mathbf{x})$: $2x_1 + e^{x_2} + 3x_1x_3 \leq 5$, where $x_1, x_2, x_3 \in \{0, 1\}$, we can find the minimal violation subsets $\mathcal{V}$ based on all possible combinations of $x_1,x_2,x_3$.

1) When $x_1 = 0, x_2 = 0, x_3 = 0$: $g(\mathbf{x})=1\leq 5$, feasible.

2) When $x_1 = 0, x_2 = 0, x_3 = 1$: $g(\mathbf{x})=1\leq 5$, feasible.

3) When $x_1 = 0, x_2 = 1, x_3 = 0$: $g(\mathbf{x})=e\leq 5$, feasible.

4) When $x_1 = 1, x_2 = 0, x_3 = 0$: $g(\mathbf{x})=3\leq 5$, feasible.

5) When $x_1 = 0, x_2 = 1, x_3 = 1$: $g(\mathbf{x})=e\leq 5$, feasible.

6) When $x_1 = 1, x_2 = 0, x_3 = 1$: $g(\mathbf{x})=6\geq 5$, violation.

7) When $x_1 = 1, x_2 = 1, x_3 = 0$: $g(\mathbf{x})=e+2\leq 5$, feasible.

8) When $x_1 = 1, x_2 = 1, x_3 = 1$: $g(\mathbf{x})=5+e\geq 5$, violation (not minimal).

Identified minimal violation subsets: $\{x_1,x_3\}$. Thus, 
\begin{equation}
    P(\mathbf{x})=\lambda (x_1x_3)
\end{equation}

Final BIP objective:
\begin{equation}
O_{\mathrm{BIP}} = f(\mathbf{x}) + \lambda(x_1x_3)
\end{equation}

\section{The hypergraph max-cut problem.}
\label{sec:hypergraphmaxcut}
The max-cut problem of a hypergraph $G=(V,E)$ involves partitioning the vertex set into two disjoint subsets such that the number of hyperedges crossing the partitioned blocks is maximized.

\textbf{PUBO Form.} The hypergraph max-cut problem on $G$ can be formulated by optimizing a PUBO objective as follows:
\begin{equation}
\mathrm{min} \quad 
O_{\mathrm{max-cut}}=\sum_{e\in E}{(1-\prod_{i\in e}{x_i}-\prod_{i\in e}{(1-x_i)})}
\label{equ:hypergrpahmaxcut}
\end{equation}
where $x_i\in\{0,1\}$ are binary decision variables.

For a simple example illustrated in Fig.~\ref{fig:hypergraph_generate}, the original hypergraph consists of three hyperedges: $\{x_1,x_2\}$, $\{x_3,x_4\}$, and $\{x_1,x_2,x_3\}$.
Thus, the max-cut objective of $G$ is to minimize $2x_1+2x_2+2x_3+x_4-3x_1x_2-x_1x_3-x_2x_3-2x_3x_4$.
BIPNN typically generates a new hypergraph structure with five hyperedges, $\{x_1,x_2\}$, $\{x_3,x_4\}$, $\{x_1,x_3\}$, and $\{x_2,x_3\}$, to solve this PUBO objective.
we found that both hypergraphs can be utilized for HyperGNN training in BIPNN framework.

\section{Datasets.}
\label{app:datasets}
\begin{table}[h]
\caption{Summary statistics of five real-world graphs: the number of vertices $|V|$, the number of edges $|E|$. Three hypergraphs: the number of vertices $|V|$, the number of hyperedges $|E|$, the size of the hypergraph $\sum_{e\in E}{|e|}$.}
\centering
\begin{tabular}{ccc|cccc}
\toprule
\textbf{Graphs} &
$|V|$ &
$|E|$ &
\textbf{Hypergraphs} &
$|V|$ &
$|E|$ &
$\sum_{e\in E}{|e|}$ \\
\midrule
BAT &
131 &
1,003 &
Primary &
242 &
12,704 &
30,729 \\
EAT &
399 &
5,993 &
High &
327 &
7,818 &
18,192 \\
UAT &
1,190 &
13,599 &
Cora &
1,330 &
1,503 &
4,599 \\
DBLP &
2,591 &
3,528 &
&
&
&
\\
CiteSeer &
3,279 &
4,552 &
&
&
&
\\
\bottomrule
\end{tabular}
%}
\label{tab:datasets_graph}
\end{table}

\end{document}